\documentclass{article}

\usepackage{makecell}
\usepackage[final]{lg}

\usepackage[utf8]{inputenc} 
\usepackage[T1]{fontenc}    
\usepackage[pdfencoding=auto, psdextra, unicode]{hyperref}       
\usepackage{url}            
\usepackage{booktabs}       
\usepackage{rotating}
\usepackage{amsfonts}       
\usepackage{amsmath}
\usepackage{nicefrac}       
\usepackage[dvipsnames]{xcolor}
\usepackage{kotex}
\usepackage{adjustbox}
\usepackage{multirow}
\usepackage{amssymb}
\usepackage{longtable}
\usepackage{comment}
\usepackage[english]{babel}
\usepackage[autostyle]{csquotes}
\usepackage{hhline}
\usepackage{tabu}
\usepackage{tablefootnote}
\usepackage{longtable}
\usepackage{graphicx}
\usepackage{wrapfig}
\usepackage{array}
\usepackage{caption}
\usepackage{tcolorbox}
\usepackage{lipsum}
\usepackage{relsize}
\usepackage{colortbl}
\usepackage{tablefootnote}
\tcbuselibrary{breakable}
\usepackage{floatpag}
\floatpagestyle{plain}
\usepackage{float}
\usepackage{threeparttable}
\usepackage{enumitem}
\setlist[enumerate,itemize]{leftmargin=1.5em}
\usepackage{tcolorbox}
\usepackage{placeins}
\usepackage{xurl}

\newtcolorbox{HeroCard}{
breakable,    
colback=black!3,
colframe=black!3,
boxrule=0pt,
arc=3mm,
left=12pt,right=12pt,top=10pt,bottom=10pt
}                                                                                                                                                                            
\definecolor{MainPurple}{HTML}{A451E4}

\newcommand{\PaperHuggingFaceURL}{https://huggingface.co/LGAI-EXAONE/EXAONE-4.5-33B}
\newcommand{\PaperGitHubURL}{https://github.com/LG-AI-EXAONE/EXAONE-4.5}                                                                                                               
  
\makeatletter
\edef\OrigRM{\rmdefault}
\edef\OrigSF{\sfdefault}
\edef\OrigTT{\ttdefault}
\renewcommand{\@maketitle}{%
\vspace*{-3.5em}   
\begin{center}
\begin{HeroCard}
    {%
      \fontfamily{put}\selectfont
      {\LARGE\bfseries \@title\par}      
  \vspace{0.5em}

  {\normalsize \@author\par}
  \vspace{0.7em}

  {\normalsize
  \setlength{\baselineskip}{1.25\baselineskip}
  \noindent 

This technical report introduces EXAONE 4.5, the first open-weight vision language model released by LG AI Research. EXAONE 4.5 is architected by integrating a dedicated visual encoder into the existing EXAONE 4.0 framework, enabling native multimodal pretraining over both visual and textual modalities. The model is trained on large-scale data with careful curation, particularly emphasizing document-centric corpora that align with LG’s strategic application domains. This targeted data design enables substantial performance gains in document understanding and related tasks, while also delivering broad improvements across general language capabilities. EXAONE 4.5 supports six languages---Korean, English, Spanish, German, Japanese, and Vietnamese---and extends context length up to 256K tokens, facilitating long-context reasoning and enterprise-scale use cases. Comparative evaluations demonstrate that EXAONE 4.5 achieves competitive performance in general benchmarks while outperforming state-of-the-art models of similar scale in document understanding and Korean contextual reasoning. As part of LG’s ongoing effort toward practical industrial deployment, EXAONE 4.5 is designed to be continuously extended with additional domains and application scenarios to advance AI for a better life.

\par}

  \vspace{1.6em}
  \noindent
  \begin{minipage}[b]{0.78\linewidth}
  {\small
    \noindent\textbf{GitHub:~} \href{\PaperGitHubURL}{\nolinkurl{\PaperGitHubURL}}\par
    \noindent\textbf{Hugging Face:~} \href{\PaperHuggingFaceURL}{\nolinkurl{\PaperHuggingFaceURL}}\par
  }
  \end{minipage}\hfill
  \begin{minipage}[b]{0.18\linewidth}
  \raggedleft
  \hspace{-8mm}\raisebox{1mm}{\includegraphics[height=5mm]{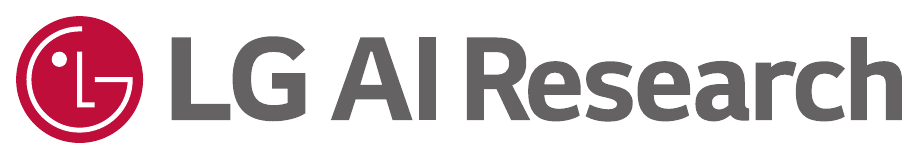}}
  \end{minipage}
}%
\end{HeroCard}
\end{center}

\@thanks
\global\let\@thanks\@empty
\setcounter{footnote}{0}
\vspace{1.25 em}
}
\makeatother

\usepackage{fourier}
\renewcommand{\rmdefault}{\OrigRM}
\renewcommand{\sfdefault}{\OrigSF}
\renewcommand{\ttdefault}{\OrigTT}

\newcommand{\model}{EXAONE 4.5 }
\newcommand{\comp}{LG~AI~Research}

\title{
\includegraphics[height=5.5mm]{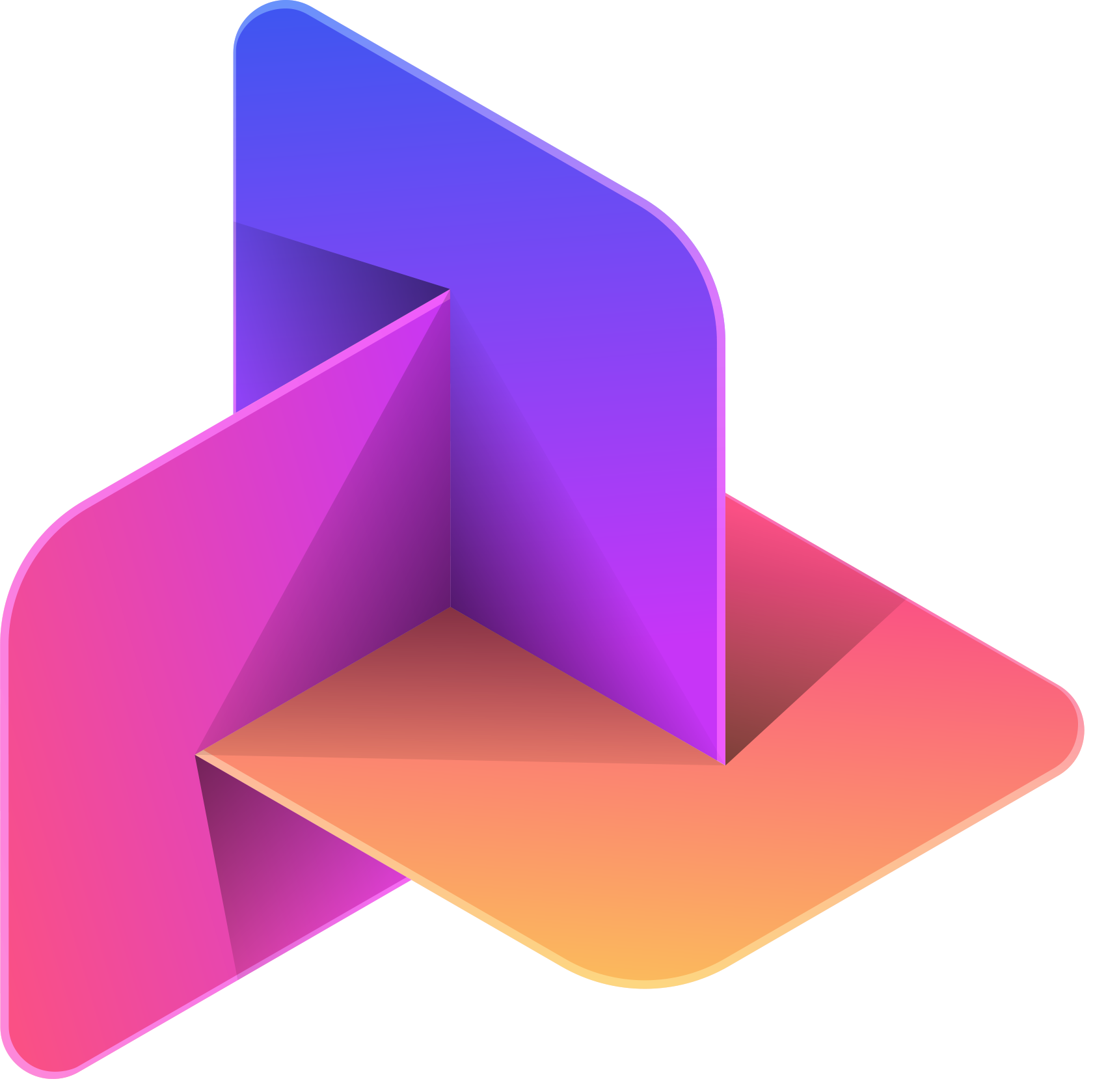}
EXAONE 4.5 Technical Report \\ \large{LG's First Open-Weight Vision-Language Model for Industrial Intelligence}
}
\author{%
  \comp\thanks{The complete list of authors who contributed to this work can be found in Appendix~\ref{appendix:contributors}.
}\\
}

\begin{document}

\maketitle

\section{Introduction}

The EXAONE foundation models have been continuously engineered to address complex and demanding challenges in real-world industrial environments. Earlier iterations, including EXAONE 3.0~\citep{an2026exaone3078binstruction} and 3.5~\citep{an2026exaone35serieslarge}, were developed as Large Language Models (LLMs) focused on integrating generative AI into practical industry applications. As the landscape shifted toward the era of Agentic AI, the necessity for advanced logical and mathematical reasoning became increasingly critical. To address this, EXAONE Deep~\citep{bae2026exaonedeepreasoningenhanced} was introduced as a specialized model optimized for reasoning tasks in domains such as mathematics, science, and coding.

Building upon these targeted capabilities, EXAONE 4.0~\citep{bae2026exaone40unifiedlarge} was designed as a hybrid LLM. It features a dual-mode architecture, employing a \textsc{Non-Reasoning} mode for general-purpose tasks---inheriting the strengths of EXAONE 3.5---and a \textsc{Reasoning} mode for tackling highly complex problems. With the release of EXAONE 4.5, we further advance this paradigm by introducing visual comprehension to these foundation models. Notably, EXAONE 4.5 represents LG's first open-weight Vision Language Model (VLM), expanding its utility beyond strictly textual boundaries.

Architecturally, EXAONE 4.5 integrates a 1.2B parameter vision encoder into the robust EXAONE 4.0 32B base model, enabling the system to seamlessly process both text and image inputs. In highly technical industrial settings, this multimodal proficiency unlocks diverse applications. For instance, in manufacturing quality control, the VLM can analyze real-time visual feeds from assembly lines to identify defects or anomalies. Similarly, in enterprise engineering and maintenance, the model can cross-reference complex visual blueprints, pipeline diagrams, and technical manuals to automate compliance checks and generate accurate operational diagnostics.

By bridging the gap between advanced language processing and visual perception, EXAONE 4.5 significantly enhances the practical problem-solving capabilities of AI in the field. Beyond its immediate utility in analyzing multimodal data, this model represents a critical milestone in our broader architectural roadmap. Ultimately, these visual and logical foundations serve as an essential stepping stone toward the realization of Vision-Language-Action (VLA) models, which will empower AI systems to autonomously interact with and operate within physical industrial environments.

\begin{figure}[!ht]
    \centering
    \includegraphics[width=\textwidth]{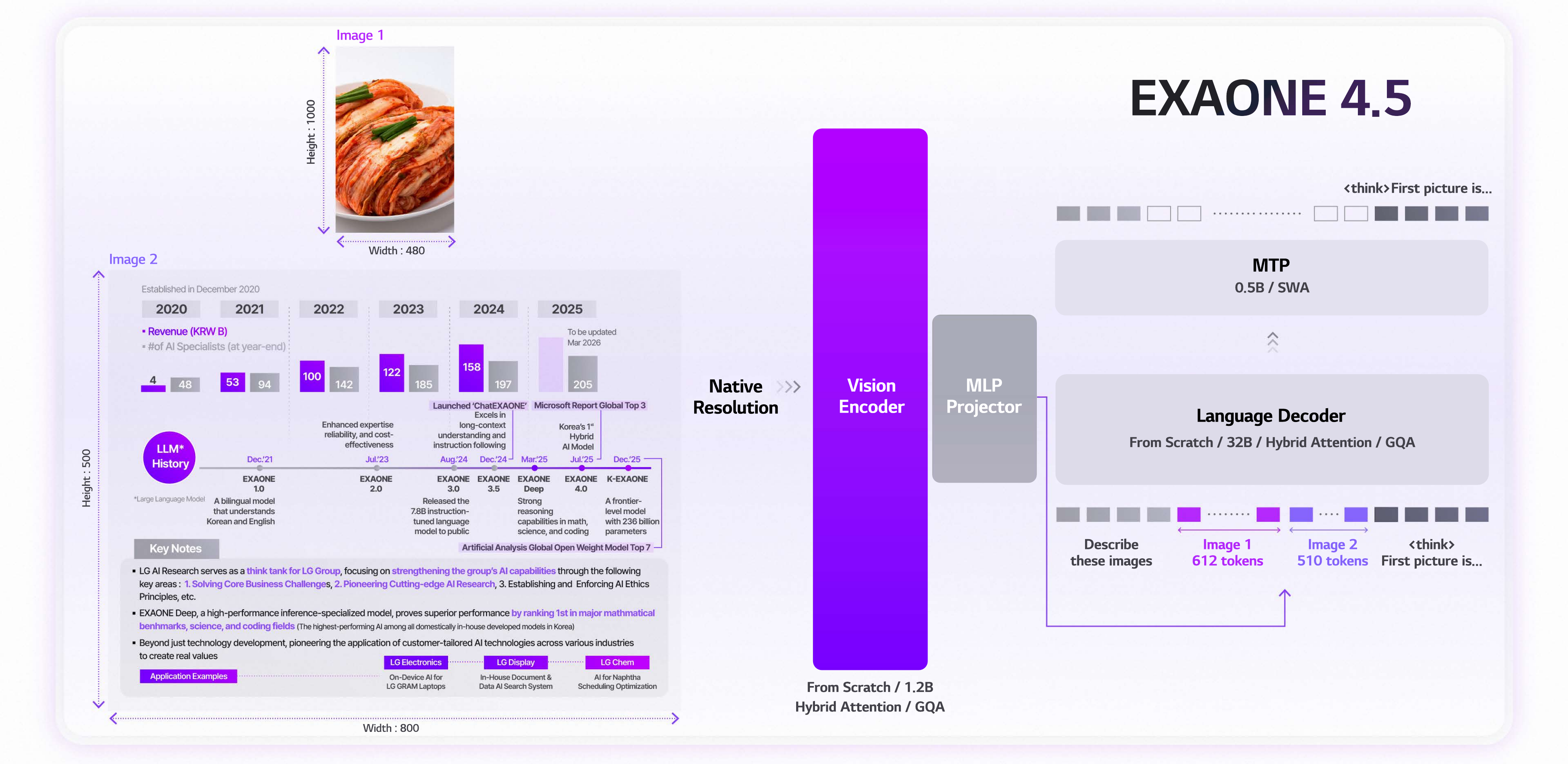}
    \caption{Overview of the EXAONE 4.5 architecture.}
    \label{fig:architecture}
\end{figure}

\section{Modeling}

\subsection{Model Configurations}

In contrast to language models, Vision-Language Models (VLMs) must process a substantial number of visual tokens originating from the vision encoder, where the token count scales with image resolution. A naive reduction in the number of visual tokens significantly degrades performance by discarding critical spatial and semantic information. To address this issue, we depart from prior approaches that rely on relatively small-scale vision encoders (e.g., ~600M parameters) and instead employ a billion-parameter-scale vision encoder. This design choice enables the model to retain rich visual representations without aggressive token truncation.

To maintain computational efficiency despite the increased model capacity, we adopt hybrid attention mechanisms alongside Grouped Query Attention (GQA)~\citep{ainslie2023gqatraininggeneralizedmultiquery}. While GQA is commonly favored for its KV cache reduction benefits in autoregressive decoders, we leverage it within the vision encoder as well. Even in the absence of KV caching, GQA provides improved end-to-end computational efficiency due to reduced attention complexity and better hardware utilization. Furthermore, GQA is widely supported and continuously optimized in modern inference frameworks, making it a pragmatic choice for deployment efficiency.

Existing vision encoders did not meet our requirements in terms of scalability and efficiency. Therefore, we trained a 1.2B-parameter vision encoder from scratch and integrated it into EXAONE 4.5. For the language backbone, we utilize the EXAONE 4.0~\citep{bae2026exaone40unifiedlarge} architecture and incorporate the Multi-Token Prediction (MTP) module~\citep{deepseekai2025deepseekv3technicalreport, gloeckle2024better} from K-EXAONE~\citep{choi2026kexaonetechnicalreport} to further improve decoding throughput.

For positional encoding, we employ 2D Rotary Positional Embedding (2D RoPE)~\citep{su2023roformerenhancedtransformerrotary} in the vision encoder to effectively capture the two-dimensional spatial structure of images, which differs fundamentally from the one-dimensional sequential nature of text. In contrast, the language model retains standard 1D RoPE to preserve compatibility with pre-trained textual positional representations, ensuring strong performance across both modalities.

This design allows us to achieve an optimal balance between performance and efficiency by mitigating the image processing bottleneck typically introduced by high-resolution inputs. Additionally, we carefully calibrate the maximum image resolution during training. While supporting ultra-high-resolution images could improve theoretical coverage, it incurs substantial computational overhead and significantly slows down training. Instead, we align the maximum resolution with commonly encountered real-world inputs, thereby optimizing resource utilization without sacrificing practical usability.

Finally, we reuse the tokenizer developed for K-EXAONE, which features significant enhancements over the EXAONE 4.0 tokenizer, particularly regarding multilingual support and Korean language processing. This ensures robust text representations across diverse linguistic contexts while maintaining consistency with prior model iterations.

\subsection{Pre-training}
\subsubsection{Overall Pipeline}
The pre-training pipeline begins with training a 1.2B-parameter vision encoder from scratch using an autoregressive objective inspired by OpenVision2~\citep{liu2025openvision}, ensuring alignment with the target architecture. This is followed by multimodal pre-training to align the encoder, vision-language merger, and LLM backbone. The pre-training process is structured into two sequential stages, designed to progressively solidify cross-modal alignment and expand the model’s representational coverage across diverse data domains.

\begin{table}[htb!]
    \centering
    \setlength{\doublerulesep}{1pt}
    \begin{tabular}{c|c|c}
        \toprule
        Stage & Stage 1 & Stage 2 \\
        \midrule
        Training Modules                  & All                       & All \\
        Image Tokens                      & 420B                      & 225B \\
        Text Tokens                       & 400B                      & 110B \\
        Sequence Length                   & 8K                        & 8K \\
        ~~Amount of computation (FLOPs)~~ & ~~$1.57 \times 10^{23}$~~ & ~~$6.43 \times 10^{22}$~~ \\
        \bottomrule
    \end{tabular}
    \vspace{2mm}
    \caption{Training configuration across pretraining stages.}
    \label{tab:training_recipe}
\end{table}

\paragraph{Stage 1: Foundational Modality Alignment}
We conduct end-to-end joint training of the vision encoder, merger, and LLM. The data mixture comprises general domain image-text pairs, interleaved image-text documents, document understanding datasets, and OCR-centric samples, specifically curated to establish robust visual-text alignment. To prevent the degradation of language modeling capabilities, text-only data, including pre- and post-training datasets from the K-EXAONE pipeline, is integrated into the training objective.

\paragraph{Stage 2: Perceptual and Knowledge Refinement}
We refine the data mixture to prioritize high-density, structured information. The curriculum reduces the proportion of general domain samples in favor of grounding, document, and OCR-centric data. Additionally, the inclusion of diverse datasets across knowledge, mathematics, and STEM domains provides the model with the necessary foundational exposure for subsequent complex multimodal tasks.

Overall, this two-stage pre-training strategy transitions the model from basic visual-text alignment to a more comprehensive understanding of structured and domain-specific multimodal data, setting a robust foundation for task-specific fine-tuning.

\subsubsection{Pre-Training Data}

\paragraph{Image Caption Data} This corpus primarily comprises Korean-English bilingual pairs. To address the brevity and noise inherent in raw web-curated captions, a synthetic captioning pipeline is employed to enhance semantic richness. The gap between general pre-training and downstream benchmark requirements is bridged by incorporating task-oriented images, including those related to mathematics, charts, diagrams, and document parsing. The pipeline ensures high quality by optimizing for visual information richness and image-text alignment. This approach prioritizes entity diversity, visual complexity, and fine-grained details while maintaining specificity and factual accuracy. To mitigate hallucinations, existing metadata serves as a reference for the synthetic generator, ensuring that descriptions are detailed and grounded in verifiable evidence.

\paragraph{Interleaved Image-Text Data} Leveraging established LLM data filtering methods~\citep{penedo2024fineweb}, a massive corpus comprising interleaved open-source~\citep{li2024omnicorpus, awadalla2024mint} and in-house resources is curated to extract high-quality multimodal web content. A lightweight text-based classifier evaluates the textual components based on educational quality scores and STEM-related relevance. This process filters out low-value web noise, enabling the strategic upsampling of documents with high information density. By preserving the natural sequence of text and images, the model improves its ability to process multimodal information within extended contexts and associate non-adjacent visual and textual cues.

\paragraph{OCR and Documents} To enhance OCR and document understanding, English and Korean datasets are constructed at the character, word, and document levels through the integration of open-source and in-house resources. Synthetic OCR images are generated with diverse backgrounds and contrastive pairs of visually confusable words. Furthermore, various document parsing tasks transform charts, tables, and documents into structured formats such as HTML, Markdown, and JSON, facilitating layout understanding and semantic structure reconstruction.

\paragraph{Grounding and Counting} A dedicated pipeline for visual grounding and counting is established to enhance spatial intelligence. Grounding data integrates high-quality open-source sets with an in-house synthetic pipeline. To ensure format consistency, all object locations are represented as bounding boxes $[x_1, y_1, x_2, y_2]$ denoting the top-left and bottom-right corners. Each coordinate is normalized by the image width or height and scaled to a range of $[0, 1000]$ following established literature~\citep{bai2025qwen3vltechnicalreport}. For counting, synthetic generation is prioritized to mitigate the noise (e.g., occlusion, crowding) found in real-world data. To counteract biases toward low count ranges and simple categories, the dataset undergoes explicit balancing across count ranges and object types, followed by iterative refinement for increased difficulty and diversity.

\paragraph{STEM and Reasoning} An integrated search-based synthesis pipeline addresses the scarcity of high-level academic content, including complex mathematical graphs, engineering, and science diagrams. This pipeline facilitates the scalable construction of datasets by retrieving and synthesizing domain-specific documents. These resources provide broad coverage of domains that are typically underrepresented in publicly available datasets. Extracted metadata informs the generation of Long Chain-of-Thought (CoT) data, which couples visual perception with deep knowledge-based reasoning. The training process follows a progressive curriculum where an initial broad filtering policy ensures general visual diversity, followed by the strategic upsampling of these specialized datasets to target remaining performance gaps.

\paragraph{Korean Specific} To address the under-representation of Korean cultural and linguistic nuances, we curate a specialized Korean multimodal corpus. Notably, datasets sourced from the Korea Tourism Organization (KTO)\footnote{https://knto.or.kr} encompass a comprehensive range of images and descriptions detailing historical and contemporary Korean culture, thereby facilitating the model's acquisition of profound cultural knowledge. Additionally, to incorporate game and IT-related content prevalent among active users, we utilized datasets from IT Donga\footnote{https://it.donga.com} and Game Donga\footnote{https://game.donga.com}, enabling the model to comprehensively cover diverse facets of modern Korean digital culture. Furthermore, to minimize hallucinations, generation is grounded in factual information, ensuring responses regarding Korea are based on verifiable evidence. For reasoning tasks, we adapt the STEM pipeline used for English and further employ an image-rendering strategy in which text-based problems are converted into high-resolution rendered images. This text-to-vision augmentation ensures the model can robustly parse and solve formatted Korean academic content.

\subsection{Context Length Extension}

The EXAONE 4.5 model supports a maximum context length of 256K tokens, achieved by integrating context extension directly into the supervised fine-tuning (SFT) stage. Unlike conventional pipelines that treat context expansion as a standalone phase, this approach leverages high-quality fine-tuning data to simultaneously refine instruction-following capabilities and long-range coherence. 

A critical advantage of this method is the stabilization provided by the 128K-capable base LLM. Initializing from a high baseline context length ensures a robust long-context prior, which minimizes optimization instability and prevents performance degradation commonly observed when scaling from shorter sequences. Furthermore, the vision encoder, already well aligned with the LLM through multimodal pretraining, contributes to stable extension in the multimodal setting.

To address the increased computational complexity of 256K sequences, we employ Context Parallelism~\citep{hao2023ringattention} to optimize memory distribution and maintain high training throughput. This integrated strategy enables stable and efficient scaling while preserving the fundamental reasoning performance of the model across extended multimodal contexts.

\subsection{Post-training}
\subsubsection{Supervised Fine-Tuning}
In the supervised fine-tuning (SFT) stage, we construct a high-quality training dataset spanning multiple domains and modalities. Rather than relying on a single unified data construction pipeline, we organize the data by domain and apply tailored curation strategies for each. The resulting dataset is designed to support a broad range of capabilities required for real-world multimodal applications, including visual understanding, language use, reasoning, and instruction following. Across all domains, we place particular emphasis on reducing noisy supervision and improving alignment between inputs, instructions, and responses.

To strengthen multimodal capabilities, we jointly train the model on text-only and vision-language data within a unified SFT framework, while integrating both \textsc{Non-Reasoning} and \textsc{Reasoning} supervision. The training dataset covers diverse visual and textual use cases, including both document-centric and general multimodal inputs, enabling the model to develop broad multimodal generalization beyond a narrow set of input types. 

Furthermore, we adopt a multi-stage training curriculum with stage-wise data organization as a coordinated training strategy. This design allows the model to progressively strengthen its overall capabilities while maintaining broad coverage and balanced development throughout training. The SFT data mixture also supports multilingual instruction following across Korean, English, Spanish, German, Japanese, and Vietnamese, enhancing the model's multimodal usability across diverse linguistic contexts.

\subsubsection{Offline Preference Optimization}
We apply offline preference optimization in a multi-stage framework, where each phase is tailored to a specific objective and integrated at different points in the training pipeline (e.g., prior to RLVR or during final preference learning).
At each stage, we adapt the optimization strategy to target particular capabilities, including OCR, chart understanding, visual recognition, dialogue, instruction following, and safety.
For example, we use $\mathcal{L}_{\text{DPO}}$~\citep{Rafael2023dpo} for vision tasks, as it provides stable optimization through a reference model.
In contrast, for text tasks, we employ $\mathcal{L}_{\text{GROUPER}}$~\citep{choi2026kexaonetechnicalreport} to more effectively leverage datasets containing multiple rejected responses.
We set $\beta = 0.1$ for DPO and fix $G = 4$ for GROUPER.

\[
\mathcal{L}_{\text{DPO}}(\theta)
= - \mathbb{E}_{(x, y^+, y^-) \sim \mathcal{D}}
\left[
\log \sigma \left(
\beta \left(
\log \frac{\pi_\theta(y^+ \mid x)}{\pi_{\text{ref}}(y^+ \mid x)}
-
\log \frac{\pi_\theta(y^- \mid x)}{\pi_{\text{ref}}(y^- \mid x)}
\right)
\right)
\right].
\]

\[
\begin{aligned}
\mathcal{L}_{\text{GROUPER}}(\theta)
&= - \mathbb{E}_{(x, y_i,...,y_G) \sim \mathcal{D}}
\left[
\frac{1}{G} \sum_{i=1}^{G}
\left(
A_{i} \cdot \exp\left(\frac{1}{|y_i|} \log \pi_{\theta}(y_i \mid x)\right)
\right)
\right], \\
\text{where } \quad
z_i
&= \frac{r_{i} - \mathrm{mean}(\{r_{j}\}_{j=1}^G)}
{\mathrm{std}(\{r_{j}\}_{j=1}^G)}, 
A_{i} = 2 \cdot
\frac{z_i - \min(\{z_j\}_{j=1}^G)}
{\max(\{z_j\}_{j=1}^G) - \min(\{z_j\}_{j=1}^G)}
- 1.
\end{aligned}
\]

\subsubsection{Reinforcement Learning}
To further enhance reasoning capability, we conduct joint multimodal reinforcement learning across both text and vision. The text data consists of mathematics, coding, knowledge, and instruction-following tasks, while the vision data encompasses general visual understanding, STEM reasoning, charts, OCR, document understanding, and multi-image settings. For reward design, we employ the same reward system as in our previous model~\citep{choi2026kexaonetechnicalreport} for text tasks and task-specific reward functions for vision tasks. For policy optimization, we adopt GRPO with the IcePop setting~\citep{shao2024deepseekmathpushinglimitsmathematical,lingteam2025stepevolvesscalingreinforcement} and apply zero-variance filtering to remove groups whose advantages are all zero. We compute advantages by subtracting the group mean reward from each sample reward, and omit standard deviation normalization to preserve training stability~\citep{deepseekai2025deepseekv32pushingfrontieropen}.

\section{Evaluation}

\subsection{Benchmarks and setup}

We evaluate EXAONE 4.5 on a broad set of vision and language benchmarks.
For vision, we group the benchmarks into four categories:
\begin{itemize}
\item \textbf{STEM / Puzzle:} \textsc{MMMU}~\citep{yue2024mmmumassivemultidisciplinemultimodal}, \textsc{MMMU-Pro}~\citep{yue2025mmmuprorobustmultidisciplinemultimodal}, \textsc{MedXpertQA-MM}~\citep{zuo2025medxpertqa}, \textsc{MATH-Vision}~\citep{wang2024measuringmultimodalmathematicalreasoning}, \textsc{MathVista}{\smaller[2]~(mini)}~\citep{lu2024mathvistaevaluatingmathematicalreasoning}, \textsc{We-Math}~\citep{qiao2024wemathdoeslargemultimodal}, \textsc{LogicVista}~\citep{xiao2024logicvistamultimodalllmlogical}, \textsc{BabyVision}~\citep{chen2026babyvision}

\item \textbf{Document Understanding:} \textsc{AI2D}~\citep{kembhavi2016diagramworthdozenimagesai2d}, \textsc{ChartQAPro}~\citep{masry2025chartqaprodiversechallengingbenchmark}, \textsc{CharXiv}{\smaller[2]~(RQ)}~\citep{wang2024charxivchartinggapsrealistic}, \textsc{OCRBench v2}~\citep{fu2025ocrbenchv2improvedbenchmark}, \textsc{OmniDocBench v1.5}~\citep{ouyang2025omnidocbenchbenchmarkingdiversepdf}

\item \textbf{General:} \textsc{MMStar}~\citep{chen2024rightwayevaluatinglargemmstar}, \textsc{BLINK}~\citep{fu2024blinkmultimodallargelanguageblink}, \textsc{HallusionBench}~\citep{guan2024hallusionbenchadvanceddiagnosticsuite}

\item \textbf{Korean:} \textsc{KMMMU}~\citep{anonymous2026kmmmu}, \textsc{K-Viscuit}~\citep{park2025kviscuitinterpretation}, \textsc{KRETA}~\citep{hwang2025kretabenchmarkkoreanreading}
\end{itemize}

For language, we use six categories:
\begin{itemize}
\item \textbf{Reasoning:} \textsc{AIME 2026}~\citep{balunovic_srimatharena_2025}, \textsc{GPQA-Diamond}~\citep{rein2023gpqagraduatelevelgoogleproofqa}, \textsc{LiveCodeBench v6}~\citep{jain2024livecodebenchholisticcontaminationfree}, \textsc{MMLU-Pro}~\citep{wang2024mmluprorobustchallengingmultitask}

\item \textbf{Agentic Tool Use:} \textsc{$\tau^2$-Bench}~\citep{barres2025tau2}

\item \textbf{Instruction Following:} \textsc{IFBench}~\citep{pyatkin2025generalizing}, \textsc{IFEval}~\citep{zhou2023instructionfollowingevaluationlargelanguage}

\item \textbf{Long Context Understanding:} \textsc{AA-LCR}~\citep{artificialanalysis2025lcr}

\item \textbf{Korean:} \textsc{KMMLU-Pro}~\citep{hong2025kmmlureduxkmmluproprofessionalkorean}, \textsc{KoBALT}~\citep{shin2025kobaltkoreanbenchmarkadvanced}
\item \textbf{Multilinguality}\footnote{We only evaluate five non-English supported languages on multilingual benchmarks: Korean (ko), German (de), Spanish (es), Japanese (ja), and Vietnamese (vi).}: \textsc{MMMLU}~\citep{hendrycks2021measuringmassivemultitasklanguage} and \textsc{WMT24++}~\citep{deutsch2025wmt24expandinglanguagecoverage}
\end{itemize}

\paragraph{Decoding settings.}
For vision benchmarks, we use a temperature of 1.0 by default and decrease it to 0.6 for the \textit{Document Understanding} and \textit{Korean} categories. We set top-$p$~\citep{Holtzman2020The} to 0.95 and apply a presence penalty of 1.5. For language benchmarks, we use a shared decoding configuration of temperature 1.0 and top-$p$ 0.95. We set the maximum generation length to 32K tokens for all vision benchmarks and 128K tokens for language benchmarks. At inference time, MTP is disabled.

For baseline models, we use officially reported scores from each model's technical report, model card, or official blog whenever available. When official scores are unavailable, we reproduce the results in our internal evaluation environment using each model's recommended inference configuration.

\subsection{Vision Benchmarks Results}
Table~\ref{tab:result_reasoning_vision} presents the results on vision benchmarks for EXAONE 4.5 and various baseline models.
EXAONE 4.5 demonstrates competitive and well-balanced performance across all four vision benchmark categories. In \textit{STEM / Puzzle}, it performs strongly on multimodal mathematical and logical reasoning. For example, it outperforms the substantially larger model, Qwen3-VL-235B, on \textsc{MathVision} (75.2 vs. 74.6) and \textsc{We-Math} (79.1 vs. 74.8), while also exceeding GPT-5 mini---a strong closed-weight model---on benchmarks such as \textsc{MMMU-Pro} (68.6 vs. 67.3) and \textsc{MathVision} (75.2 vs. 71.9).

In the \textit{Document Understanding} category, EXAONE 4.5 also demonstrates solid capabilities on diagram- and document-centric tasks. For example, it surpasses Qwen3-VL-235B on \textsc{CharXiv}{\smaller[2]~(RQ)} (71.7 vs. 66.1) and outperforms GPT-5 mini on representative benchmarks including \textsc{AI2D} (89.0 vs. 88.2) and \textsc{OmniDocBench}{\smaller[2]~(v1.5)} (81.2 vs. 77.0).
Performance is also stable on \textit{General} benchmarks; notably, EXAONE 4.5 outperforms Qwen3-VL-235B on \textsc{BLINK} (68.7 vs. 67.1).
Overall, these results indicate that EXAONE 4.5 provides robust vision-language reasoning performance across a broad range of evaluation scenarios.

\begin{table*}[!tp]
\centering
\small
\setlength{\tabcolsep}{4pt}
\caption{The main evaluation results of EXAONE 4.5 \textsc{Reasoning} mode on vision benchmarks.}
\label{tab:result_reasoning_vision}
\begin{threeparttable}
\resizebox{1.0\textwidth}{!}{
\begin{tabular}{@{}lccccc@{}}
\toprule
\multicolumn{1}{l|}{}
& \multicolumn{1}{c|}{\makecell{\textbf{~~EXAONE 4.5 33B~~} \\ \smaller[1]~\textbf{(\textsc{Reasoning})}}} & 
                      \makecell{GPT-5~mini~\citep{singh2025openaigpt5card} \\ {\smaller[1]~(\textsc{Reasoning: high})}} & \makecell{Qwen3-VL-32B~\citep{bai2025qwen3vltechnicalreport}\\Thinking} & 
                      \makecell{Qwen3-VL-235B-A22B~\citep{bai2025qwen3vltechnicalreport}\\Thinking} & 
                      \makecell{Qwen3.5-27B~\citep{qwen3.5} \\ {\smaller[1]~(\textsc{Reasoning})}} \\
\midrule

\multicolumn{1}{l|}{Architecture}        & \multicolumn{1}{c|}{Dense} & - & Dense & MoE  & Dense \\
\multicolumn{1}{l|}{\# Total Params}     & \multicolumn{1}{c|}{33B}   & - & 33B   & 236B & 27B \\
\multicolumn{1}{l|}{\# Activated Params} & \multicolumn{1}{c|}{33B}   & - & 33B   & 23B  & 27B \\
\midrule

\rowcolor[rgb]{0.9,0.9,0.9}\multicolumn{6}{c}{\textit{STEM / Puzzle}} \\
\midrule

\multicolumn{1}{l|}{\textsc{MMMU}}                          & \multicolumn{1}{c|}{78.7} & 79.0 & 78.1 & 80.6 & 82.3 \\ 
\multicolumn{1}{l|}{\textsc{MMMU-Pro}}                      & \multicolumn{1}{c|}{68.6} & 67.3 & 68.1 & 69.3 & 75.0 \\ 
\multicolumn{1}{l|}{\textsc{MedXpertQA-MM}}                 & \multicolumn{1}{c|}{42.1} & 34.4 & 41.6 & 47.6 & 62.4 \\
\multicolumn{1}{l|}{\textsc{MATH-Vision}}                   & \multicolumn{1}{c|}{75.2} & 71.9 & 70.2 & 74.6 & 86.0 \\ 
\multicolumn{1}{l|}{\textsc{MathVista}{\smaller[2]~(mini)}} & \multicolumn{1}{c|}{85.0} & 79.1 & 85.9 & 85.8 & 87.8 \\ 
\multicolumn{1}{l|}{\textsc{We-Math}}                       & \multicolumn{1}{c|}{79.1} & 70.3 & 71.6 & 74.8 & 84.0 \\ 
\multicolumn{1}{l|}{\textsc{LogicVista}}                    & \multicolumn{1}{c|}{73.8} & 70.3 & 70.9 & 72.2 & 77.0 \\ 
\multicolumn{1}{l|}{\textsc{BabyVision}}                    & \multicolumn{1}{c|}{18.8} & 20.9 & 17.4 & 22.2 & 44.6 \\ 
\midrule

\rowcolor[rgb]{0.9,0.9,0.9}\multicolumn{6}{c}{\textit{Document Understanding}} \\
\midrule

\multicolumn{1}{l|}{\textsc{AI2D}}                             & \multicolumn{1}{c|}{89.0} & 88.2 & 88.9 & 89.2 & 92.9 \\
\multicolumn{1}{l|}{\textsc{ChartQAPro}}                       & \multicolumn{1}{c|}{62.2} & 60.9 & 61.4 & 61.2 & 66.8 \\ 
\multicolumn{1}{l|}{\textsc{CharXiv}{\smaller[2]~(RQ)}}        & \multicolumn{1}{c|}{71.7} & 68.6 & 65.2 & 66.1 & 79.5 \\
\multicolumn{1}{l|}{\textsc{OCRBench} v2}       & \multicolumn{1}{c|}{63.2} & 55.8 & 68.4 & 66.8 & 67.3 \\
\multicolumn{1}{l|}{\textsc{OmniDocBench v1.5}} & \multicolumn{1}{c|}{81.2} & 77.0 & 83.1 & 84.5 & 88.9 \\
\midrule

\rowcolor[rgb]{0.9,0.9,0.9}\multicolumn{6}{c}{\textit{General}} \\
\midrule

\multicolumn{1}{l|}{\textsc{MMStar}}         & \multicolumn{1}{c|}{74.9} & 74.1 & 79.4 & 78.7 & 81.0 \\
\multicolumn{1}{l|}{\textsc{BLINK}}          & \multicolumn{1}{c|}{68.8} & 67.7 & 68.5 & 67.1 & 71.6 \\ 
\multicolumn{1}{l|}{\textsc{HallusionBench}} & \multicolumn{1}{c|}{63.7} & 63.2 & 67.4 & 66.7 & 70.0 \\ 
\midrule

\rowcolor[rgb]{0.9,0.9,0.9}\multicolumn{6}{c}{\textit{Korean}} \\
\midrule

\multicolumn{1}{l|}{\textsc{KMMMU}}     & \multicolumn{1}{c|}{42.7} & 42.6 & 37.8 & 42.1 & 51.7 \\ 
\multicolumn{1}{l|}{\textsc{K-Viscuit}} & \multicolumn{1}{c|}{80.1} & 78.5 & 78.5 & 83.9 & 84.0 \\ 
\multicolumn{1}{l|}{\textsc{KRETA}}     & \multicolumn{1}{c|}{91.9} & 94.8 & 90.3 & 92.8 & 96.5 \\ 
\bottomrule

\end{tabular}
}
\end{threeparttable}
\end{table*}
\begin{table*}[!tp]
\centering
\small
\setlength{\tabcolsep}{4pt}
\caption{The main evaluation results of EXAONE 4.5 \textsc{Reasoning} mode on language benchmarks.}
\label{tab:result_reasoning_language}
\begin{threeparttable}
\resizebox{1.0\textwidth}{!}{
\begin{tabular}{@{}lccccc@{}}
\toprule
\multicolumn{1}{l|}{} 
& \multicolumn{1}{c|}{\makecell{\textbf{~~EXAONE 4.5 33B~~} \\ \smaller[1]~\textbf{(\textsc{Reasoning})}}} & 
                      \makecell{K-EXAONE-236B-A23B~\citep{choi2026kexaonetechnicalreport}\\ {\smaller[1]~(\textsc{Reasoning})}} &
                      \makecell{GPT-5~mini~\citep{singh2025openaigpt5card} \\ {\smaller[1]~(\textsc{Reasoning: high})}} & 
                      \makecell{Qwen3-VL-235B-A22B~\citep{bai2025qwen3vltechnicalreport} \\Thinking} & 
                      \makecell{Qwen3.5-27B~\citep{qwen3.5} \\ {\smaller[1]~(\textsc{Reasoning})}} \\
\midrule

\multicolumn{1}{l|}{Architecture}        & \multicolumn{1}{c|}{Dense} & MoE  & - & MoE  & Dense \\
\multicolumn{1}{l|}{\# Total Params}     & \multicolumn{1}{c|}{33B}   & 236B & - & 236B & 27B \\
\multicolumn{1}{l|}{\# Activated Params} & \multicolumn{1}{c|}{32B}   & 23B  & - & 22B  & 27B \\
\midrule

\rowcolor[rgb]{0.9,0.9,0.9}\multicolumn{6}{c}{\textit{Reasoning}} \\
\midrule

\multicolumn{1}{l|}{\textsc{AIME 2026}}        & \multicolumn{1}{c|}{92.6} & 92.2 & 92.4 & 89.4 & 93.2 \\ 
\multicolumn{1}{l|}{\textsc{GPQA-Diamond}}     & \multicolumn{1}{c|}{80.5} & 79.1 & 82.3 & 77.1 & 85.5 \\ 
\multicolumn{1}{l|}{\textsc{LiveCodeBench v6}} & \multicolumn{1}{c|}{81.4} & 80.7 & 78.1 & 70.1 & 80.7 \\ 
\multicolumn{1}{l|}{\textsc{MMLU-Pro}}         & \multicolumn{1}{c|}{83.3} & 83.8 & 83.3 & 83.8 & 86.1 \\ 

\midrule

\rowcolor[rgb]{0.9,0.9,0.9}\multicolumn{6}{c}{\textit{Agentic Tool Use}} \\
\midrule

\multicolumn{1}{l|}{\textsc{$\tau^2$-Bench} {\smaller[2](\textsc{retail})}}  & \multicolumn{1}{c|}{77.9} & 78.6 & 78.3 & 67.0 & 84.7 \\ 
\multicolumn{1}{l|}{\textsc{$\tau^2$-Bench} {\smaller[2](\textsc{airline})}} & \multicolumn{1}{c|}{56.5} & 60.4 & 60.0 & 62.0 & 67.5 \\ 
\multicolumn{1}{l|}{\textsc{$\tau^2$-Bench} {\smaller[2](\textsc{telecom})}} & \multicolumn{1}{c|}{73.0} & 73.5 & 74.1 & 44.7 & 99.3 \\ 
\midrule

\rowcolor[rgb]{0.9,0.9,0.9}\multicolumn{6}{c}{\textit{Instruction Following}} \\
\midrule

\multicolumn{1}{l|}{\textsc{IFBench}}          & \multicolumn{1}{c|}{62.6} & 67.3 & 74.0 & 59.2 & 76.5 \\ 
\multicolumn{1}{l|}{\textsc{IFEval}}           & \multicolumn{1}{c|}{89.6} & 89.7 & 92.8 & 88.2 & 95.0 \\ 

\midrule

\rowcolor[rgb]{0.9,0.9,0.9}\multicolumn{6}{c}{\textit{Long Context Understanding}} \\
\midrule

\multicolumn{1}{l|}{\textsc{AA-LCR}}           & \multicolumn{1}{c|}{50.6} & 53.5 & 68.0 & 58.7 & 67.3 \\ 
\midrule

\rowcolor[rgb]{0.9,0.9,0.9}\multicolumn{6}{c}{\textit{Korean}} \\
\midrule

\multicolumn{1}{l|}{\textsc{KMMLU-Pro}}        & \multicolumn{1}{c|}{67.6} & 67.3 & 72.5 & 71.1 & 73.0 \\ 
\multicolumn{1}{l|}{\textsc{KoBALT}}           & \multicolumn{1}{c|}{52.1} & 61.8 & 63.6 & 51.1 & 54.9 \\
\midrule
\rowcolor[rgb]{0.9,0.9,0.9}\multicolumn{6}{c}{\textit{Multilinguality}} \\
\midrule

\multicolumn{1}{l|}{\textsc{MMMLU}~{\smaller[2]~(ko,de,es,ja)}}        & \multicolumn{1}{c|}{85.4} & 85.7 & 88.1 & 86.8 & 88.9 \\ 
\multicolumn{1}{l|}{\textsc{WMT24++}~{\smaller[2]~(ko,de,es,ja,vi)}}           & \multicolumn{1}{c|}{91.5} & 90.5 & 96.7 & 94.4 & 94.7 \\ 
\bottomrule

\end{tabular}
}
\end{threeparttable}
\end{table*}

\subsection{Language Benchmarks Results}
As shown in Table~\ref{tab:result_reasoning_language}, EXAONE 4.5 exhibits its greatest strengths in core reasoning and coding tasks within the language benchmarks. In the \textit{Reasoning} category, it achieves the best score on \textsc{LiveCodeBench v6}, outperforming all other compared models. It also delivers particularly strong results on \textsc{AIME 2026}, ranking second overall and first among all baselines except Qwen3.5-27B. These results indicate that EXAONE 4.5 is especially competitive in code-related evaluations as well as mathematical reasoning.

Beyond core reasoning, EXAONE 4.5 remains competitive on agentic tool-use and instruction-following benchmarks. Using a weighted average over \textsc{$\tau^2$-Bench} subsets, EXAONE 4.5 achieves 72.0, substantially outperforming Qwen3-VL-235B-A22B (57.0). It also outperforms Qwen3-VL-235B-A22B across the \textit{Instruction Following} category, achieving higher scores on both \textsc{IFBench} (62.6 vs. 59.2) and \textsc{IFEval} (89.6 vs. 88.2).
Overall, these results indicate that EXAONE 4.5 delivers robust language reasoning performance across a broad range of evaluation settings.

\section{Limitations} \label{Limitations}

\model{}models, like all existing multimodal models, have certain limitations and may occasionally generate inappropriate responses. The multimodal model generates responses based on the output probability of tokens, and it is determined during learning from training data. While we make every effort to exclude personal, harmful, and biased information from the training data, some problematic content may still be included, potentially leading to undesirable responses. Please note that the text generated by \model{}models does not reflect the views of LG AI Research.

\begin{itemize}
    \item Inappropriate answers may be generated, which contain personal, harmful or other inappropriate information.
    \item Biased responses may be generated, which are associated with age, gender, race, and so on.
    \item The generated responses rely heavily on statistics from the training data, which can result in the generation of semantically or syntactically incorrect sentences.
    \item Since the models do not reflect the latest information, the responses may be false or contradictory.
\end{itemize}
	
LG AI Research strives to reduce potential risks that may arise from \model{}models. Users are not allowed to engage in any malicious activities (e.g., keying in illegal information) that may induce the creation of inappropriate outputs violating LG AI's ethical principles when using \model{}models.

\section{Deployment}

Section~\ref{appendix:license} in the Appendix provides license information for using the \model models. Understanding the license information is essential for the legal utilization of the language model.

\section{Conclusion}

This report introduces EXAONE 4.5, advancing the EXAONE foundation model series and marking LG's first open-weight VLM. By integrating a custom-built 1.2B-parameter vision encoder with the EXAONE 4.0 32B language model, we have successfully bridged the gap between advanced reasoning and visual comprehension.

Our architectural and training innovations---including the application of GQA in the vision encoder, 2D RoPE, and the MTP module---ensure an optimal balance between computational efficiency and model capacity. Through a rigorous, multi-stage pre-training pipeline, EXAONE 4.5 has acquired robust capabilities in processing complex multimodal data, ranging from general image-text pairs to highly structured OCR, STEM diagrams, and specialized Korean cultural and academic content. Furthermore, by embedding context extension directly into the supervised fine-tuning stage, we achieved a stable and effective context length of 256K tokens without compromising cross-modal alignment. Combined with our tailored preference optimization and multimodal RL, the model demonstrates strong reasoning and instruction-following abilities.

Extensive evaluations across a comprehensive suite of benchmarks validate the efficacy of these design choices. EXAONE 4.5 exhibits highly competitive, state-of-the-art performance across diverse vision and language tasks. It frequently outperforms substantially larger or closed-weight models in complex domains such as mathematical reasoning, document parsing, and agentic tool use, proving its robust capability to handle intricate, real-world problems.

Ultimately, EXAONE 4.5 serves as a powerful problem-solving engine tailored for demanding industrial environments. Beyond its immediate utility in analyzing multimodal data, this model establishes a critical foundation for the future development of VLA models. By releasing EXAONE 4.5 as an open-weight model, we aim to accelerate community-driven research and foster the next generation of AI systems.

\newpage

\appendix

\clearpage
\section{Contributors}
\label{appendix:contributors}
All authors are listed in alphabetical order by last name.

\paragraph{Core Contributors}
Eunbi~Choi, Kibong~Choi, Sehyun~Chun, Seokhee~Hong, Junwon~Hwang, Hyojin~Jeon, Ahra~Jo, Hyunjik~Jo, Yeonsik~Jo, Joonkee~Kim, Seonghwan~Kim, Soyeon~Kim, Sunkyoung~Kim, Yireun~Kim, Yongil~Kim, Changhun~Lee, Haeju~Lee, Jinsik~Lee, Kyungmin~Lee, Sangha~Park, Kwangrok~Ryoo, Minju~Seo, Sejong~Yang, Heuiyeen~Yeen

\paragraph{Contributors}
Hwan~Chang, Stanley~Jungkyu~Choi, Yejin~Choi, Kyubeen~Han, Joonwon~Jang, Kijeong~Jeon, \mbox{Geunyeong~Jeong}, Gerrard~Jeongwon~Jo, Jiyeon~Jung, Daeseong Kim, Dohoon~Kim, Dohyun~Kim, Hyunseo~Kim, Minu~Kim, Myoungshin~Kim, Youchul~Kim, Byungoh~Ko, Christopher~Lee, Edward~Hwayoung~Lee, Honglak~Lee, Jiyoung~Lee, Sangeun~Lee, Seungwon~Lim, Woohyung~Lim, Jueun~Mun, Jaewoo~Park, Jimin~Park, Jinho~Park, \mbox{Yongmin~Park}, Wooseok~Seo, Yongwoo~Song, Sihyuk~Yi, Kyungjae~Yoo, Sangyeon~Yoon

\newpage

\section{Model License}
\label{appendix:license}

\textbf{EXAONE AI Model License Agreement 1.2 - NC} \\
\\
This License Agreement (“Agreement”) is entered into between you (“Licensee”) and LG Management Development Institute Co., Ltd. (“Licensor”), governing the use of the EXAONE AI Model (“Model”). By downloading, installing, copying, or using the Model, you agree to comply with and be bound by the terms of this Agreement. If you do not agree to all the terms, you must not download, install, copy, or use the Model. This Agreement constitutes a binding legal agreement between the Licensee and Licensor. \\
\\
\\
\textbf{1. Definitions} \\
\\
\textbf{1.1 Model:} The artificial intelligence model provided by Licensor, which includes any software, algorithms, machine learning models, or related components supplied by Licensor. This definition extends to encompass all updates, enhancements, improvements, bug fixes, patches, or other modifications that may be provided by Licensor from time to time, whether automatically or manually implemented. \\
\\
\textbf{1.2 Derivatives:} Any modifications, alterations, enhancements, improvements, adaptations, or derivative works of the Model created by Licensee or any third party. This includes changes made to the Model's architecture, parameters, data processing methods, or any other aspect of the Model that results in a modification of its functionality or output. \\ 
\\
\textbf{1.3 Output:} Any data, results, content, predictions, analyses, insights, or other materials generated by the Model or Derivatives, regardless of whether they are in their original form or have been further processed or modified by the Licensee. This includes, but is not limited to, textual or numerical produced directly or indirectly through the use of the Model. \\
\\
\textbf{1.4 Licensor:} LG Management Development Institute Co., Ltd., the owner, developer, and provider of the EXAONE AI Model. The Licensor holds all rights, title, and interest in the Model and is responsible for granting licenses to use the Model under the terms specified in this Agreement. \\
\\
\textbf{1.5 Licensee:} The individual, organization, corporation, academic institution, government agency, or other entity using or intending to use the Model under the terms and conditions of this Agreement. The Licensee is responsible for ensuring compliance with the Agreement by all authorized users who access or utilize the Model on behalf of the Licensee. \\
\\
\\
\textbf{2. License Grant} \\ 
\\
\textbf{2.1 Grant of License:} Subject to the terms and conditions outlined in this Agreement, the Licensor hereby grants the Licensee a limited, non-exclusive, non-transferable, worldwide, and revocable license to: \\
\\
a. Access, download, install, and use the Model solely for research and educational purposes. This includes evaluation, testing, academic research, experimentation, learning, teaching, training and participation in competitions, provided that such participation is in a non-commercial context. Notwithstanding Section 3.1, the Licensee may only provide the Model or Derivatives for a competition if no commercial license is granted to the competition organizer or any third party. \\
\\
b. Publicly disclose research results and findings derived from the use of the Model or Derivatives, including publishing papers or presentations. \\
\\
c. Modify the Model and create Derivatives based on the Model, provided that such modifications and Derivatives are used exclusively for research and educational purposes. The Licensee may conduct experiments, perform analyses, and apply custom modifications to the Model to explore its capabilities and performance under various scenarios. If the Model is modified, the modified Model must include “EXAONE” at the beginning of its name. \\
\\
d. Distribute the Model and Derivatives in each case with a copy of this Agreement. \\
\newpage
\textbf{2.2 Scope of License:} The license granted herein does not authorize the Licensee to use the Model for any purpose not explicitly permitted under this Agreement. Any use beyond the scope of this license, including any commercial application or external distribution, is strictly prohibited unless explicitly agreed upon in writing by the Licensor. \\
\\
\\
\textbf{3. Restrictions}\label{textbf:Restrictions} \\
\\
\textbf{3.1 Commercial Use:} The Licensee is expressly prohibited from using the Model, Derivatives, or Output for any commercial purposes, including but not limited to, developing or deploying products, services, or applications that generate revenue, whether directly or indirectly. Any commercial exploitation of the Model or its derivatives requires a separate commercial license agreement with the Licensor. Furthermore, the Licensee shall not use the Model, Derivatives or Output to develop or improve any models that compete with the Licensor’s models. \\
\\
\textbf{3.2 Reverse Engineering:} The Licensee shall not decompile, disassemble, reverse engineer, or attempt to derive the source code, underlying ideas, algorithms, or structure of the Model, except to the extent that such activities are expressly permitted by applicable law. Any attempt to bypass or circumvent technological protection measures applied to the Model is strictly prohibited. \\
\\
\textbf{3.3 Unlawful Use:} The Licensee shall not use the Model and Derivatives for any illegal, fraudulent, or unauthorized activities, nor for any purpose that violates applicable laws or regulations. This includes but is not limited to the creation, distribution, or dissemination of malicious, deceptive, or unlawful content. \\
\\
\textbf{3.4 Ethical Use:} The Licensee shall ensure that the Model or Derivatives is used in an ethical and responsible manner, adhering to the following guidelines: \\
\\
a. The Model and Derivatives shall not be used to generate, propagate, or amplify false, misleading, or harmful information, including fake news, misinformation, or disinformation. \\
\\
b. The Model and Derivatives shall not be employed to create, distribute, or promote content that is discriminatory, harassing, defamatory, abusive, or otherwise offensive to individuals or groups based on race, gender, sexual orientation, religion, nationality, or other protected characteristics. \\
\\
c. The Model and Derivatives shall not infringe on the rights of others, including intellectual property rights, privacy rights, or any other rights recognized by law. The Licensee shall obtain all necessary permissions and consents before using the Model and Derivatives in a manner that may impact the rights of third parties. \\
\\
d. The Model and Derivatives shall not be used in a way that causes harm, whether physical, mental, emotional, or financial, to individuals, organizations, or communities. The Licensee shall take all reasonable measures to prevent misuse or abuse of the Model and Derivatives that could result in harm or injury. \\
\\
\\
\textbf{4. Ownership} \\
\\
\textbf{4.1 Intellectual Property:} All rights, title, and interest in and to the Model, including any modifications, Derivatives, and associated documentation, are and shall remain the exclusive property of the Licensor. The Licensee acknowledges that this Agreement does not transfer any ownership rights to the Licensee. All trademarks, service marks, and logos associated with the Model are the property of the Licensor. \\
\\
\textbf{4.2 Output:} Licensor claims no rights in Output. Licensee is solely responsible for the Output and its use. \\
\\
\textbf{4.3 Attribution:} In any publication or presentation of results obtained using the Model, the Licensee shall provide appropriate attribution to the Licensor, citing the Model's name and version, along with any relevant documentation or references specified by the Licensor. \\
\newpage
\textbf{5. No Warranty} \\
\\
\textbf{5.1 “As-Is” Basis:} The Model, Derivatives, and Output are provided on an “as-is” and “as-available” basis, without any warranties or representations of any kind, whether express, implied, or statutory. The Licensor disclaims all warranties, including but not limited to, implied warranties of merchantability, fitness for a particular purpose, accuracy, reliability, non-infringement, or any warranty arising from the course of dealing or usage of trade. \\
\\
\textbf{5.2 Performance and Reliability:} The Licensor does not warrant or guarantee that the Model, Derivatives or Output will meet the Licensee’s requirements, that the operation of the Model, Derivatives or Output will be uninterrupted or error-free, or that defects in the Model will be corrected. The Licensee acknowledges that the use of the Model, Derivatives or Output is at its own risk and that the Model, Derivatives or Output may contain bugs, errors, or other limitations. \\
\\
\textbf{5.3 No Endorsement:} The Licensor does not endorse, approve, or certify any results, conclusions, or recommendations derived from the use of the Model. The Licensee is solely responsible for evaluating the accuracy, reliability, and suitability of the Model for its intended purposes. \\
\\
\\
\textbf{6. Limitation of Liability} \\
\\
\textbf{6.1 No Liability for Damages:} To the fullest extent permitted by applicable law, in no event shall the Licensor be liable for any special, incidental, indirect, consequential, exemplary, or punitive damages, including but not limited to, damages for loss of business profits, business interruption, loss of business information, loss of data, or any other pecuniary or non-pecuniary loss arising out of or in connection with the use or inability to use the Model, Derivatives or any Output, even if the Licensor has been advised of the possibility of such damages. \\
\\
\textbf{6.2 Indemnification:} The Licensee agrees to indemnify, defend, and hold harmless the Licensor, its affiliates, officers, directors, employees, and agents from and against any claims, liabilities, damages, losses, costs, or expenses (including reasonable attorneys' fees) arising out of or related to the Licensee's use of the Model, any Derivatives, or any Output, including any violation of this Agreement or applicable laws. \\
\\
\\
\textbf{7. Termination} \\
\\
\textbf{7.1 Termination by Licensor:} The Licensor reserves the right to terminate this Agreement and revoke the Licensee’s rights to use the Model at any time, with or without cause, and without prior notice if the Licensee breaches any of the terms or conditions of this Agreement. Termination shall be effective immediately upon notice. \\
\\
\textbf{7.2 Effect of Termination:} Upon termination of this Agreement, the Licensee must immediately cease all use of the Model and Derivatives and destroy all copies of the Model and Derivatives in its possession or control, including any backup or archival copies. The Licensee shall certify in writing to the Licensor that such destruction has been completed. \\
\\
\textbf{7.3 Survival:} The provisions of this Agreement that by their nature should survive termination, including but not limited to, Sections 4 (Ownership), 5 (No Warranty), 6 (Limitation of Liability), and this Section 7 (Termination), shall continue to apply after termination. \\
\\
\\
\textbf{8. Governing Law} \\
\\
\textbf{8.1 Governing Law:} This Agreement shall be governed by and construed in accordance with the laws of the Republic of Korea, without regard to its conflict of laws principles. \\
\\
\textbf{8.2 Arbitration:} Any disputes, controversies, or claims arising out of or relating to this Agreement, including its existence, validity, interpretation, performance, breach, or termination, shall be referred to and finally resolved by arbitration administered by the Korean Commercial Arbitration Board (KCAB) in accordance with the International Arbitration Rules of the Korean Commercial Arbitration Board in force at the time of the commencement of the arbitration. The seat of arbitration shall be Seoul, Republic of Korea. The tribunal shall consist of one arbitrator. The language of the arbitration shall be English. \\
\\
\textbf{9. Alterations} \\
\\
\textbf{9.1 Modifications:} The Licensor reserves the right to modify or amend this Agreement at any time, in its sole discretion. Any modifications will be effective upon posting the updated Agreement on the Licensor’s website or through other means of communication. The Licensee is responsible for reviewing the Agreement periodically for changes. Continued use of the Model after any modifications have been made constitutes acceptance of the revised Agreement. \\
\\
\textbf{9.2 Entire Agreement:} This Agreement constitutes the entire agreement between the Licensee and Licensor concerning the subject matter hereof and supersedes all prior or contemporaneous oral or written agreements, representations, or understandings. Any terms or conditions of any purchase order or other document submitted by the Licensee in connection with the Model that are in addition to, different from, or inconsistent with the terms and conditions of this Agreement are not binding on the Licensor and are void. \\
\\
By downloading, installing, or using the EXAONE AI Model, the Licensee acknowledges that it has read, understood, and agrees to be bound by the terms and conditions of this Agreement. \\







\newpage
\bibliographystyle{plain} 
\bibliography{refs} 


\end{document}